\documentclass[11pt]{article}
\usepackage[margin=1in]{geometry}
\usepackage{amsmath,amssymb,booktabs,graphicx,microtype}
\usepackage[colorlinks=true,allcolors=blue]{hyperref}
\usepackage{natbib}
\usepackage{xcolor}

\newcommand{\diag}{(t,t)}

\title{Where a Model Sends Its Own Repeated Token\\
{\large A vocabulary-wide destination map, its measured robustness envelope,\\and the estimand it
replaced}}

\author{Nicol\'as Vera Z\'u\~niga\\
Independent Researcher, Chile\\
\texttt{nicovera@quetru.cl}
}
\date{}

\begin{document}
\maketitle

\begin{abstract}
Black-box model identification is an established problem with established solutions, and they work
by scoring a model's response to natural-language prompts. One line of work does feed models a
degenerate input --- their own token, repeated --- but in order to find a failure mode rather than an
identity. We take that same input and ask a different question of it: not whether the model gets
stuck, but where it goes when it does not. For each token $t$ in a vocabulary, read
$\arg\max p(\cdot \mid t, t)$ in a single forward pass. The resulting object is a map defined on the
whole vocabulary, and it has two halves. The first --- which tokens are fixed points of that map ---
is partially anticipated, and we report it as a \emph{failed} estimand: the natural distance on it is
$83\%$ explained by set cardinality alone, it separates a corpus manipulation by two bits in $3471$
against a precision floor of zero, and it attributes families at $0.5833$. The second half, where the
map sends tokens that are \emph{not} fixed points, is not recorded in prior work; the one paper that
had those tokens logged them as a zero. Pairing the comparison on the source token removes the
cardinality confound by construction ($r$ from $0.9128$ to $-0.0932$) and attributes model families
at $0.8333$ --- twelve models scored against a pool of nineteen --- with a chance rate of
$0.1389$, across seven tokenizer groups and several corpora. Two nulls clear it: frequency-matched destinations agree at $0.1429$, independent marginals
at $0.0798$. Family predicts agreement better than tokenizer ($0.2031$ against $0.1205$), and
recurrent architectures cluster, at balanced accuracy $1.0$ against a majority rate of $0.7895$ and
$0.90$ once each model's dominant destination is excluded, which is the figure we stand behind. We
measure the robustness envelope rather than assume it: $8$-bit weight rounding moves the map less
than deduplicating the training corpus does ($0.9004$ against $0.6353$, on one support), $4$-bit destroys it
($0.0098$, and $0.1812$ at the granularity deployment actually uses, so the failure is not an
artefact of coarse quantization), and the numeric precision floor varies by model from $0.201$ to
$0.9778$, so every robustness statement here is per-model. All estimands, thresholds and kill conditions were registered
before the data, and the failed one is reported at the same length as the surviving one.
\end{abstract}

\section{Introduction}\label{sec:intro}

Deciding which model is behind an endpoint is a solved-enough problem to have a literature. Model
equality testing, provenance testing and single-shot fingerprinting all work, and they work by
sending natural-language prompts and scoring what comes back
\citep{mpt2025,stemma2026,onetoken2026}. We are not proposing a better version of that. We are asking
a different question: what does a model do when the input is not language at all?

The input we use is the simplest degenerate one available --- a token followed by itself --- and the
reading is a single forward pass. Two quantities fall out. Whether the state $\diag$ reproduces
itself under the model's own $\arg\max$, and, when it does not, \emph{which token it produces
instead}. The first is a bit per token; the second is a destination per token. Neither requires
generation, sampling, a temperature, or a prompt.

\paragraph{What is already known, and what we therefore do not claim.} The probe shape is not ours.
\citet{hammouri2025nonhalting} feed models their own repeated tokens and formalise the
temperature-zero condition under which the state reproduces itself. The argmax map over token space
as a dynamical object is also not ours: our own earlier work \citep{veraz2026probes} identifies it and
reports its per-model structure. Section~\ref{sec:delta} states both deltas in full, because a
novelty claim that has to be reconstructed by the reader is not a novelty claim.

\paragraph{What this paper contributes.} A map defined on the \emph{whole vocabulary} rather than
sampled from it, valued in \emph{token identities} rather than counts, read as decoded strings so
that models with different tokenizers are comparable, and characterised by a measured robustness
envelope rather than an assumed one.

\paragraph{And a negative result we report at length.} We registered the set of fixed points as the
feature first. It fails, for a reason worth printing: the natural distance between two such sets is
dominated by how large they are rather than by which tokens they contain. We report that in
\S\ref{sec:e1} at the same length as the result that works, because a paper that shows only its
surviving estimand gives the reader no way to judge whether it was the first one tried.

\section{Setup}\label{sec:setup}

\paragraph{The measurement.} For a model $M$ and a token $t$, one forward pass on the two-token input
$\diag$ gives $\arg\max_x p_M(x \mid t, t)$. Define the \emph{self-continuation bit} as whether that
argmax is $t$ itself, and the \emph{destination} as the argmax when it is not. We also store the
logit margin between the winner and the runner-up. Nothing is sampled and nothing is generated;
there is no temperature and no prompt. The measurement is deterministic, and we assert that rather
than assume it: every cell is re-measured and required to be bit-for-bit identical before it is
written.

\paragraph{The probe set, frozen before any model was loaded.} Token identifiers are not comparable
across tokenizers, so the index is a list of \emph{strings}: the $2000$ most frequent case-sensitive
words of $2000$ Pile documents, bare and space-prefixed, plus the printable ASCII block both ways and
a fixed whitespace list --- $4090$ candidates, hashed and committed before any measurement. Each
model resolves them against its own tokenizer, and the comparison set is the intersection: strings
encoding to exactly one token everywhere.

\paragraph{A rule that inverts, and confusing the two manufactures findings.} Within one model, token
identifiers are authoritative and decoded strings are not, because two identifiers can print alike.
\emph{Across} models the rule reverses: identifiers are meaningless and the string is the only
bridge. We key sources by the frozen strings and compare destinations as decoded strings, and we
report the hazards of that bridge --- leading spaces, non-ASCII, empty decodings --- rather than
assume them away.

\paragraph{Cohort.} Nineteen models spanning Pythia, GPT-Neo, RWKV, Mamba, Llama, Gemma, Qwen, OLMo,
Falcon, StableLM and SmolLM, with seven distinct tokenizer groups. Twelve are Pile-trained; the rest
are not, so corpus varies. It is not \emph{controlled}: this design can say that a result is not
confined to one corpus, and cannot attribute anything to corpus.

\section{Related work, and what separates this from it}\label{sec:delta}

\paragraph{The probe is not new; the index set is.}
The degenerate diagonal input and its fixed-point reading are published.
\citet{hammouri2025nonhalting} feed a model its own token repeated, formalise the temperature-zero
condition --- at $\tau = 0$, a fixed point $x$ of $f$ with $f(x_1,x_2,x_3) = x_1,x_2,x_3$ yields an
output that never halts --- and tabulate, for $100$ randomly chosen words across five aligned models,
how many repetitions each word needs before the model stops emitting its end-of-string token. They
further observe that a fixed point found in a base model transfers to models derived from it, which
is the same invariance our attribution rests on. We claim none of this.

Three things separate the present work from theirs, and only the third is a claim about novelty
rather than scope. First, the measurement: theirs is behavioural and requires generation, since
non-halting is defined by what the model does over many steps, while ours is a single forward pass
per token and generates nothing. Second, the sample: one hundred words against five aligned models
--- two open-weight, three served only through an API --- against an exhaustive sweep of $50$k-token
vocabularies across nineteen models. Third, and this is the delta: their table enters $0$ wherever
the model halts normally, and discards what it emitted instead. Those cells are where our
measurement begins. We ask which token the model produced in place of the repetition, read that
answer across the whole vocabulary, and compare it between models as a decoded string. Every result
below rests on the contents of their zeros.

\paragraph{What our own earlier work established, and what it did not.}
The argmax map over token space is not introduced here. Paper~1 \citep{veraz2026probes} identifies it
as the mechanism behind the transition it reports and states the per-model contrast directly: for
\texttt{pythia-410m} the map sends $18$ of $24$ random starts to the newline token, a genuine fixed
point, while \texttt{gpt2-medium} has no such point and wanders to $11$ distinct endpoints. It also
shows the map's behaviour is a property of its domain rather than its parameters, one prepended token
moving the frozen fraction from $74.4\%$ to $24.1\%$. The object, the mechanism, and the per-model
contrast are therefore already ours and already in print, and this paper claims none of them.

What paper~1 measures is a trajectory census: twenty-four random starts, iterated, with the number of
distinct endpoints as the readout. It reports how many places trajectories land, and never which
tokens are fixed points, because twenty-four starts cannot enumerate a vocabulary. This paper inverts
that. It evaluates every token of every model's vocabulary exactly once, keeps identity rather than
count, and keeps it for the tokens that are \emph{not} fixed points as much as for those that are.
The difference is not one of scale. A census of endpoints and a map defined on the whole vocabulary
are different objects, and only the second supports the comparison the results turn on --- which
tokens, agreeing between which models.

\paragraph{The claim, stated narrowly.}
What is new here is a vocabulary-wide, set-valued destination map: for every token in a model's
vocabulary, where its own two-token diagonal state is sent by the argmax of the model's conditional,
decoded to a string so that models with different tokenizers are comparable at all. Not the probe,
which is \citet{hammouri2025nonhalting}'s. Not the fixed-point framing, which is theirs and
paper~1's. Not the agreement statistic, which is the established estimand of the model-provenance
line \citep{stemma2026,mpt2025}. Not the leave-one-out family-attribution protocol, which is already
published for a single-shot probe battery \citep{onetoken2026}. The index set, and what is recorded
at each of its elements.

\section{E1: the set of fixed points, and why we abandoned it}\label{sec:e1}

We registered the self-continuation set as the feature: the bit vector over the shared probe
intersection, compared between models by Hamming distance. The intersection over twelve Pile-trained
models is $3471$ strings, of which $1684$ vary across the cohort, so the estimand had room.

It does not work, and the numbers are worth printing.

\begin{table}[t]\centering\small
\begin{tabular}{lrr}
\toprule
comparison & Hamming & robust at $\tau = 1$ \\
\midrule
\texttt{pythia-410m} vs \texttt{-deduped} (a corpus manipulation) & $2$ & $1$ \\
vs \texttt{gpt-neo-125m} & $276$ & $137$ \\
vs \texttt{rwkv-4-430m} & $440$ & $198$ \\
vs \texttt{mamba-370m} & $661$ & $316$ \\
\midrule
the same weights at two numeric precisions & $0$ & $0$ \\
\bottomrule
\end{tabular}
\caption{The registered estimand separates families and does not separate a corpus manipulation from
numeric noise. Source: F183, \texttt{results/selfcont\_verdict.json}.}
\label{tab:e1}
\end{table}

\paragraph{Two bits against a floor of zero.} The pair that differs only in training-corpus
deduplication is separated by $2$ bits out of $3471$, and by \emph{zero} at the strictest margin
threshold, while the far comparisons hold hundreds. The kill condition we registered --- that the
decisive distance must exceed the precision floor --- passes, because the floor is $0$. Its own
registered corollary is what carries: a floor of exactly zero makes the test weak, since any nonzero
distance clears it.

\paragraph{And the distance is mostly set size.} Across the $66$ pairs, the Hamming distance
correlates with the sum of the two set sizes at $r = 0.9128$. Some $83\%$ of what looked like an
identity comparison is cardinality. The reason is arithmetic rather than subtle: for sparse sets,
Hamming \emph{is} $|A| + |B| - 2|A \cap B|$. The consequence is visible in the errors ---
\texttt{rwkv-4-169m} is nearer \texttt{pythia-1b} at $69$ than its own sibling
\texttt{rwkv-4-430m} at $412$, while their overlap coefficients are $0.6923$ and $0.7297$. Family
attribution on this estimand is $0.5833$.

We had registered an anti-vacuity gate on this estimand, and it did not fire, because it gates a
different degeneracy: coordinates that are constant across the cohort, which contribute zero to a
Hamming distance. That reasoning is correct and incomplete, and the gate passing is what made the
defect survive inspection.

\section{E2: where the map sends the tokens that do not stay}\label{sec:e2}

For a source token whose bit is $0$, the argmax is the destination. Comparing two models by the
\emph{agreement rate over shared source tokens} --- paired on the source, destinations compared as
decoded strings --- removes the cardinality confound by construction rather than by adjustment:
$r$ falls from $0.9128$ to $-0.0932$.

\begin{table}[t]\centering\small
\begin{tabular}{lrr}
\toprule
comparison & agreement & disagreement vs the floor \\
\midrule
the same weights at two numeric precisions & $0.7127$ & --- \\
\texttt{pythia-410m} vs \texttt{-deduped} & $0.6355$ & $1.27\times$ \\
vs \texttt{gpt-neo-125m} & $0.3902$ & $2.12\times$ \\
vs \texttt{mamba-370m} & $0.3493$ & $2.26\times$ \\
vs \texttt{rwkv-4-430m} & $0.3393$ & $2.30\times$ \\
\bottomrule
\end{tabular}
\caption{The destination map resolves the corpus manipulation above numeric noise, where the fixed-point
set could not. Source: F185, \texttt{results/escape\_destinations.json}.}
\label{tab:e2}
\end{table}

\paragraph{Both nulls clear, and one of them is the test rather than a control.} Roughly $50{,}000$ of
a Pythia's tokens do not self-continue, and their destination is whatever that model's generic
high-probability continuation is. Two models agreeing about that would be a fact about token
frequency, not about the models. Against destinations redrawn inside their own frequency band the
decisive pair agrees at $0.1429$; under independent marginals it would agree at $0.0798$. The
observed $0.6355$ is not a frequency effect.

\section{E3: nineteen models, seven tokenizers, several corpora}\label{sec:e3}

The twelve-model result has two confounds we could only disclose: both attribution errors were Mamba
models landing on RWKV, so what was recovered might be architecture class rather than family; and
eleven of the twelve shared a GPT-NeoX or GPT-2 vocabulary, so family and tokenizer were the same
variable. Widening the cohort makes both testable. All three questions were registered with kill
conditions before the new cells existed.

\begin{table}[t]\centering\small
\begin{tabular}{llrr}
\toprule
question & statistic & value & baseline \\
\midrule
family attribution & rank-1, leave-one-out & $0.8333$ & $0.1389$ chance \\
architecture class & balanced accuracy & $1.0$ & $0.7895$ majority \\
family vs.\ tokenizer & lift in mean agreement & $0.2031$ & $0.1205$ tokenizer \\
\bottomrule
\end{tabular}
\caption{$19$ models, intersection $3355$ strings, seven tokenizer groups, corpus not fixed. Source:
F188, \texttt{results/escape\_widening.json}.}
\label{tab:e3}
\end{table}

\paragraph{Family attribution is not principally a tokenizer effect.} With seven tokenizer groups
rather than a near-uniform vocabulary, agreement is better predicted by shared family than by shared
tokenizer. This is the question the narrower cohort could not ask.

\paragraph{Recurrent architectures cluster, and here is exactly how much that is worth.} Every model's
nearest neighbour is in its own architecture class. But the recurrent class is four models from two
families --- the same four whose errors generated the hypothesis --- because the hybrid
Mamba/attention model we registered as the informative test case failed to load. The widening added
seven transformer distractors and no new recurrent model. And when each model's single most common
destination is excluded, the perfect separation becomes $0.90$: above the base rate, not perfect. The
honest figure is $0.90$.

\paragraph{What the widening did not widen.} No new family reached two members, so the twelve models
scored for attribution are the same twelve as before; what grew is the pool of distractors they are
scored against. The chance rate falls accordingly, from $0.2273$ to $0.1389$.

\section{The robustness envelope, measured}\label{sec:robust}

A fingerprint that only works at full precision on the machine that measured it is not a fingerprint.
We registered the perturbations before running them.

\paragraph{Quantization.} Weight-only symmetric per-channel round-to-nearest over every linear
layer, with embeddings and norms untouched. At $8$ bits the destination map agrees with its
full-precision self at $0.9004$ --- \textbf{less movement than deduplicating the training corpus
produces} ($0.6353$). Both are computed on the same $3355$-string support, $3352$ and $3351$
escaping sources respectively; per-model probe sets differ, and a number from one may not be set
beside a number from another. At $4$ bits it collapses to $0.0098$, and the fingerprint is scoped
to full precision. That negative is located rather than merely bounded. Per-output-channel rounding
is the coarsest granularity possible and nothing deployed uses it, so we repeated the measurement
over groups of $128$ input channels --- the granularity GPTQ, AWQ and bitsandbytes all quantize at.
Grouped $4$-bit reaches $0.1812$: an order of magnitude better, and still short of $0.6353$ by a
factor of three and a half. Granularity is not what kills it. What remains untested is calibration,
the half those methods add on top of grouping, and we do not implement it.

\paragraph{The fixed-point set does not survive $4$ bits either, and its natural statistic hides
that.} At $4$ bits \texttt{pythia-410m}'s bit vector differs from its full-precision self on $8$
positions out of $3471$, which reads as near-perfect robustness. The model has $8$ self-continuing
tokens in that index and keeps $0$ of them. We report kept fractions, never Hamming counts, for the
reason \S\ref{sec:e1} gives.

\paragraph{Numeric precision, and why it is reported per model.}
Reading the same weights at \texttt{bfloat16} instead of \texttt{float32} changes the destination on
$2\%$ of sources for one model and on $80\%$ for another. There is no single precision floor for this
instrument: they run from $0.201$ to $0.9778$ across eight models, so every robustness statement here
is per-model. Within \texttt{pythia} the floor rises monotonically with scale, $0.201$ at $70$M to
$0.8989$ at $1$B, but that ordering does not survive leaving the family --- \texttt{gpt-neo-125m} is
the smaller model and the more robust, at $0.9569$. In \texttt{float32} the
measurement is invariant to how tokens are grouped into batches ($0$ of $18254$ argmaxes change); in
\texttt{bfloat16} it is not ($163$ of $3683$), so part of any \texttt{bfloat16} floor is
irreproducibility rather than precision.

\paragraph{A confidence threshold fixes the floor and destroys the signal with it.} Restricting to
sources where the winner beats the runner-up by a margin raises the precision floor toward $1.0$, so
the flipped destinations are near-ties. But it removes exactly the low-confidence sources that
carried the discrimination: at the threshold we registered as primary, the corpus pair sits at
$0.9981$ against a floor of $0.9983$ --- indistinguishable from numeric noise. Signal and noise live
in the same place. An intermediate threshold looks much better, and it is not a result:
surveying every model's floor across the ladder, \emph{all eight} sit at or above $0.99$ at that
threshold. The ratio has no stable denominator anywhere in the cohort, so the rung is not merely
unreplicated --- it is unresolvable on this estimand, and we retire it.

\section{Limits}\label{sec:limits}

The causal gap first. This is an observational comparison of checkpoints. It cannot say \emph{why}
two models disagree about where a token goes, and the design separates neither architecture nor
training schedule nor hyperparameters.

\textbf{Family is confounded with tokenizer, less than before but not removed.} Seven tokenizer
groups over nineteen models is enough to show family predicts agreement better; it is not enough to
say tokenizer contributes nothing.

\textbf{The architecture result rests on four models from two families}, the same four whose errors
generated the hypothesis, and its perfect separation is $0.90$ once the dominant destination is
excluded. A hybrid architecture would be the
informative test; ours could not be loaded --- a weight-tying incompatibility between the checkpoint
and the modelling code, not a capacity limit --- and the test is owed rather than answered.

\textbf{Corpus varies but is not controlled}, so no claim here is a corpus claim.

\textbf{The index set is a shared-string intersection, and that bounds every cross-lingual
extension.} Common support shrinks as a cohort diversifies, and it shrinks fastest in the direction
diversity comes from: adding a model with a $30$k non-Latin vocabulary took our intersection from
$3471$ strings to $152$. Pairwise intersections with per-pair normalisation, or byte-level anchors,
would remove this; neither is used here.

\textbf{Quantization beyond weight rounding is untested.} Activation quantization, real serving
stacks and deployed quantized checkpoints are all outside what we measured.

\textbf{Instance identification is not attempted and cannot be.} The measurement is deterministic, so
repeated measurement of one checkpoint is bit-identical and the test that would license an instance
claim cannot fail. What is reported is family attribution.

\section{Conclusion}

A model's response to its own repeated token, read across the whole vocabulary and kept as token
identities rather than counts, attributes model families at $0.8333$ against $0.1389$ chance, with
family outweighing tokenizer, and survives $8$-bit weight rounding better than it survives a change
of training corpus. The half of that object which was already published --- which tokens are fixed
points --- does not support the same comparison, and we report why at length: the natural distance on
it measures how many there are rather than which ones. The distinction is the paper. What a model
does when it does \emph{not} repeat itself turns out to carry more about which model it is than
whether it repeats itself at all.

\bibliographystyle{plainnat}
\bibliography{refs}

\end{document}